# Self-Taught Support Vector Machine

Parvin Razzaghi[1,2]

[1] Department of Computer Science and Information Technology, Institute for Advanced Studies in Basic Sciences (IASBS), Zanjan, Iran.

[2] School of Computer Science, Institute for Research in Fundamental Sciences (IPM), Tehran, Iran.

p.razzaghi@iasbs.ac.ir

**ABSTRACT**

In this paper, a new approach for classification of target task using limited labeled target data as well as enormous unlabeled source data is proposed which is called self-taught learning. The target and source data can be drawn from different distributions. In the previous approaches, covariate shift assumption is considered where the marginal distributions $p(x)$ change over domains and the conditional distributions $p(y|x)$ remain the same. In our approach, we propose a new objective function which simultaneously learns a common space $\Im(.)$ where the conditional distributions over domains $p(\Im(x)|y)$ remain the same and learns robust SVM classifiers for target task using both source and target data in the new representation. Hence, in the proposed objective function, the hidden label of the source data is also incorporated. We applied the proposed approach on Caltech-256, MSRC+LMO datasets and compared the performance of our algorithm to the available competing methods. Our method has a superior performance to the successful existing algorithms.

# 1 INTRODUCTION

Transfer learning is a research area in machine learning, which states that the knowledge gained from a source task, can be transferred to a related target task. It is shown that knowledge transfer, if it is applied correctly, would improve the performance of learning. Knowledge transfer can be beneficial in many machine learning areas such as classification and regression. In this paper, knowledge transfer in a classification task is considered. Up to now, many approaches have been introduced in transfer learning [1]. Similar to [1], we categorize transfer learning methods based on different situations between the source and target datasets into four categories: 1) labeled data is available in the target and source domains. In this case, it is similar to the multi-task learning setting. 2) labeled data is available in the target domain but there is no labeled data in the source domain. This setting is similar to self-taught learning. The first and second categories are referred to as inductive transfer learning. 3) No labeled data is available in the target domain but there is labeled data in the source domain which is stated as transductive transfer learning [2, 3], and 4) there is no labeled data in the target and source domain which is referred to as unsupervised learning.

In this work, we propose a new approach in inductive transfer learning where the source data is unlabeled. This setting of inductive transfer learning is called self-taught learning. Suppose that the target data $(x^t, y^t)$ are distributed according to $P_T(x, y)$ and the source data $(x^s, y^s)$ are sampled from $P_S(x, y)$. In inductive transfer learning, it is assumed that the two datasets are sampled from the two different distributions $P_T(x, y) \neq P_S(x, y)$. Also, in self-taught learning setting, the number of the target data is typically much smaller that of the source data. Hence, a model learned only by target data does not usually provide a good prediction quality. As a result, the source data is employed to achieve a good prediction quality of a target model. Inductive transfer learning approaches, based on what knowledge is transferred, are divided into two categories. In the first category, it is assumed that the joint distributions of both domains are similar to some degree $P_T(x, y) \approx P_S(x, y)$. Hence, target data model should be similar to the source data model. In the second category, source data is used to find a common domain in which target and source data have a similar distribution. In this case, it is assumed that the two datasets are sampled from different distributions such that the conditional distribution over labels $P_T(y|x) = P_S(y|x)$ is assumed to be the same, but the marginal distribution $P_T(x) \neq P_S(x)$ is considered different in target and source datasets [4-6]. Hence, in the referred works, the target and source samples are transformed to a new representation in which they have the same marginal distribution and then the target samples in the new space are only employed in minimizing target classifier's objective function. In this work, we will investigate situations where conditional distribution over $x$ is different ($P_T(x|y) \neq P_S(x|y)$). Moreover, this situation occurs more in self-taught learning setting where we have enormous unlabeled data without any assumption.

In this paper, we propose a new unified framework for self-taught learning. In our approach, we simultaneously learn a transformed space $\Im(.)$ where the conditional distribution over domains remains the same ($P_T(\Im(x)|y) \approx P_S(\Im(x)|y)$) and learns robust SVM classifiers for target task using both source and target data in the transformed space. It should be noted that the source data typically helps boost the performance of the target classifier significantly. In our approach, we propose a new objective function which utilizes the target data and the knowledge of the source data and its hidden labels in learning the target task. The training phase of our approach is an iterative algorithm which has two main steps. The first step is done by fixing the latent labels of the source data and finding a new space in which the conditional distribution of the target and source samples are similar and also the target and source data are linearly separable. In the second step, the domain is assumed to be fixed and then the latent labels of the source data are updated by optimizing an energy function. The proposed algorithm is called Self-Taught SVM (STSVM) (see Figure 1).

In our work, the source samples are assumed to be unlabeled and are not a part of the test set. In comparison, in the transductive transfer learning, the learning algorithm knows exactly which samples it will be evaluated on after training. In other words, the unlabeled target samples are test samples.

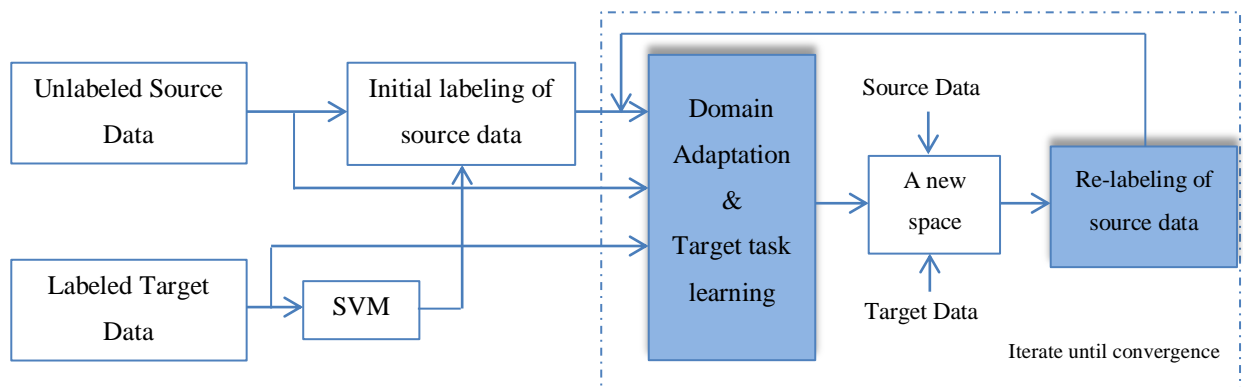

Figure 1- The overall scheme of the proposed approach

The main contributions of this paper are as follows: 1) proposing a new domain adaptation criterion which considers the class conditional distribution. Hence, our approach finds a new space in which both target and source data have similar conditional distribution over samples, 2) Incorporating the hidden labels of the source data as well as labels of target data in optimizing target classifier model in the new space, and 3) Finding a new space and the target classifier model are done simultaneously.

The rest of the paper is organized as follows. In section 2, related works are given. The problem formulation is presented in Section 3. Section 4 describes the optimization algorithm of the proposed energy function. In section 5, the results of applying our proposed approach to some category of best well known Caltech256 and MSRC-LMO datasets are given and discussed. Concluding remarks are given in Section 6.

## 2 RELATED WORKS

In this section, we review the most prominent approaches in inductive and transductive transfer learning.

*Inductive Transfer Learning.* Inductive transfer learning approaches, based on what knowledge is transferred, are divided into two categories. The first category uses the pre-trained classifier models of the source data to learn a classifier model for target data. In [7], a new model for target data is obtained such that it is encouraged to be close to one of pre-trained models, whereas Tommasi et al. [8], found a new model for target data such that it should be close to a set of pre-trained models. In [8], an SVM-based method for learning object categories from few examples is presented. In their work, the knowledge of the old learned models is transferred to the model of the target data by closing target model to the weighted mean of pre-trained models. To do this, a convex optimization problem which minimizes an estimate of the generalization error is solved. Ablavsky et al. [9] jointly trained two SVMs where the two learned models share the support vectors.

The second category uses the source data to find a domain in which the target and source data have a similar distribution. Raina et al. [10], used the unlabeled source data to find a set of basis vectors using sparse coding. Then, the target samples are represented by these basis vectors. Finally, a classifier is learned by applying a supervised learning algorithm using the target samples in the new representation. One of the challenges of this work is that the basis vectors only span over the source domain; therefore, it might not provide a good representation for the target domain. Wang et al. [11] introduced a robust loss function to learn to automatically select the optimal dictionary basis vectors. Moreover, the supervision information contained in the labeled data is also incorporated. In [12], a Feature Replication (FR) method is proposed to use the source data to augment features for cross-domain learning. The augmented features are then employed to construct a kernel function for Support Vector Machine (SVM) training. Duan et al. [13] introduced a new approach which is called DTSVM. In their work, the target and source domains are transferred to a space where the difference between data distributions is reduced. To do this, they employed means of samples from the two domains in the Reproducing Kernel Hilbert Space (RKHS). Then, they provide a unified framework to simultaneously learn an optimal kernel function and a robust SVM classifier. Li et al. [14] used the structural information in data to learn a dictionary. They learned a dictionary across the auxiliary and target domains, with expressive codings for the samples in the target domain. Also, to consider subspace structures, a low-rank constraint is introduced into the coding objective to characterize the structure of the given target set.

Chen et al. [15] introduced a new co-training approach for domain adaptation which is called CODA. In CODA, the source and target samples are all labeled. CODA is an iterative algorithm in which in each iteration, $c$ most confident predictions on unlabeled test samples are moved to the labeled samples for the next iteration. Also, in each iteration, the subset of features are selected for domain adaptation such that they are similarly predictive across the training set and the unlabeled set, instead of across the two domains. As more unlabeled test target instances are added to the training set, the target specific features become compatible across the two sets and consequently are included in the predictor.

*Transductive Transfer Learning.* Transductive learning approaches can be categorized into three categories. In the first category, the target task is learned on source labeled data by reweighting them such that the source instances which are close to the target instances are more important. In [16], a weight function is asymptotically defined to be the ratio of the density function of the covariate in the target samples to the source samples. Bickel et al. [17] learns the weights of source samples discriminatively. In the second category, a new representation space is found where the source and target data have the same distribution [18-22]. Li et al. [22] introduced a new approach named Topic Correlation Analysis (TCA) for cross-domain text classification. They extract both the shared and the domain-specific latent topics to transfer the knowledge between domains. The shared topics are investigated to identify the correlations between the domain-specific topics from different domains. If the domain-specific topics are related to many shared topics, then they are semantically correlated and should be mapped to each other. Finally, each sample is represented in a new shared space spanned by both the shared and the mapped domain-specific topics. Theoretical frameworks for domain adaptation are also introduced. In the theoretical works [18, 23], the risk of the target model is upper-bounded jointly by the model's risk on the source distribution, the divergence between the marginal distributions, and a non-estimable term related to the ability to adapt in the current space. Gang et al. [21], introduced a new approach in transductive transfer learning in which they extract conditional transferable components $T(x)$ whose conditional distribution is invariant $p(T(x)|y)$ after proper location-scale (LS) transformations. As far as we know, this is the first approach in inductive transfer learning which investigates situations where the conditional distribution over $x$ is different. There is a significant difference between our work and [21]. In their work, to consider the conditional probability in the domain adaptation term, the MMD measure is modified such that the density ratio of source labels is incorporated. However, in our approach, the MMD measure is modified such that it encodes the difference between the conditional mean of projected source and target samples in the embedded system. To do this, the source and target labels are incorporated explicitly. Moreover, in our approach, the discriminative information contained in the source and target samples are also considered. In [19], the first PAC-Bayesian analysis for domain adaptation is introduced. They proposed a pseudo-metric which evaluates the domain divergence according to the disagreement of posterior distribution over the hypothesis class. Germain et al. [20], derived a new upper bound for PAC-Bayesian domain adaptation where the distributions' divergence controls the trade-off between the source error measure and the target voters disagreement. In the third category, the target task model is modified by incorporating the pseudo-labeled information of the target data. In other words, the self-labeling of the unlabeled data is done [24, 25]. Self-labeling methods try to include unlabeled target domain samples in the training process by initializing labels of target samples and then iteratively refine their labels. Bruzzone & Marconcini [26] assume that training and test data are drawn from a different distribution. Hence, they incorporate domain adaptation technique in SVM formulation which is called DASVM. It is an iterative-based algorithm in which unlabeled target data are labeled according to the current decision function, and then a subset of the unlabeled target samples and their estimated labels is iteratively selected and moved into the training set to learn the target task. Morvant [27] presented a new framework called PV-MinCq that generalizes the non-adaptive MinCq algorithm [28]. In their formulation, self-labels of the target samples are incorporated in energy function.

## 3 PROBLEM FORMULATION

In this section, a problem formulation is given. Assume a labeled dataset from the target domain is given by $T = \{x_i^t, y_i^t\}_{i=1}^{N_t}$ in which the feature space and class space are represented by $x^t \in \chi$ and $y^t \in \{0,1\}$ respectively.

Target samples are drawn *i.i.d.* from the joint distribution $P_T$. Also, a data set from the source domain is provided which is shown by $S = \{x_i^s\}_{i=1}^{N_s}$ where the source data is unlabeled. Source samples are drawn *i.i.d.* from the marginal distribution $D_S$. It should be noted that, in this research, there is no assumption that the data from the source and target domains share the same distribution or class labels. The main goal is to use the target data and the source data to learn a function $f : \chi \rightarrow \{0,1\}$ that maps the feature space of target domain into the class space. In other words, we propose an approach to use the source data to learn a model to better classify the target data. In this formulation, $N_t$ and $N_s$ denote the number of the target and the source data, respectively. Ben-David et al. [18] have proved the Theorem 1, in which a bound is provided under the assumption that there exists a hypothesis in hypothesis class $H$ that performs well on both the source and the target domains. Let $h : \chi \rightarrow \{0,1\}$ be a hypothesis function, then $R_{P_T}(h)$, the expected target error of $h$ over $P_T$, is defined as:

$$R_{P_T}(h) = \underset{(x^t, y^t) \sim P_T}{E} L_{0-1}(h(x^t), y^t) \tag{1}$$

where $L_{0-1}$ is a zero-one loss function which maps to zero if the arguments are the same and to 1 otherwise. The expected source error $R_{P_S}(h)$ over $P_S$ is defined in the same way. Also, the expected target disagreement of $h$ and $h'$ is defined as follows:

$$R_{D_T}(h, h') = \underset{x^t \sim D_T}{E} L_{0-1}(h(x^t), h'(x^t)) \tag{2}$$

where $D_T$ is the marginal distribution of the target samples. The expected source disagreement $R_{D_S}(h, h')$ over $D_S$ is similarly defined.

**Theorem 1.** Let $H$ be a (symmetric) hypothesis class. Then, we have [18]:

$$\forall h \in H, \; R_{P_T}(h) \leq R_{P_S}(h) + \frac{1}{2} d_{H\Delta H}(D_S, D_T) + R_{P_S}(h^*) + R_{P_T}(h^*) \tag{3}$$

Where $\frac{1}{2} d_{H\Delta H}(D_S, D_T)$ is the $H\Delta H - $ distance between the marginal $D_S$ and $D_T$ which is defined as follows:

$$\frac{1}{2} d_{H\Delta H}(D_S, D_T) = \underset{(h,h') \in H^2}{\sup} | R_{D_T}(h, h') - R_{D_S}(h, h') | \tag{4}$$

Also, $h^* = \arg\min_{h \in H}(R_{P_S}(h) + R_{P_T}(h))$ where $h^*$ is the best hypothesis in hypothesis class $H$.

Under Theorem 1, the risk of the target model is upper-bounded jointly by the model's risk on the source and target data and the divergence between the marginal distributions. In our approach, we propose to design a self-taught learning algorithm which utilizes Theorem 1. To do so, we minimize the next objective function:

$$\min_{h, \Im} \hat{R}_T(h_\Im) + \hat{R}_S(h_\Im) + \hat{d}(\Im(T), \Im(S)) \tag{5}$$

where $h_\Im(.) = h(\Im(.))$, hence $\hat{R}_T(h_\Im)$ and $\hat{R}_S(h_\Im)$ represent the empirical model's risk on the transformed target and source samples under transformation function $\Im(.)$, respectively. The third term $\hat{d}(\Im(T), \Im(S))$ which is called domain adaptation term measures the difference of *conditional distribution* between the transformed source and target samples. In the following, each term is demonstrated in detail. Then, in the next section, the optimization process of Equation 5 is explained.

Figure 2 demonstrates a setting in which the domain difference is substantially large. We randomly generate two clouds of samples as target samples, each of which contains 5 samples (see Figure 2-a). We simply treat the red samples as positive target samples and green samples as negative training samples. Also, two clouds of 200 samples are generated as source samples (see Figure 2-b). It should be noted that source samples are unlabeled. In Figure 2-c, the target and source samples are shown on the same plot. Obviously, the distributions of both target and source domains are significantly different. In Figure 2-d, the source and target samples are transformed to a domain such that the distance between the means of target and source samples is close. In Figure 2-e, the source and target samples are transformed to a domain such that they have the same marginal distribution. By utilizing our proposed approach, the source and target samples are transferred to a new domain such that they have the same conditional distribution (see Figure 2-g). In this case, the source samples in the transformed space could be effectively used to improve learning on a target task. Therefore, Figure 2-g is a good intuition for the effectiveness of our approach for domain adaptation.

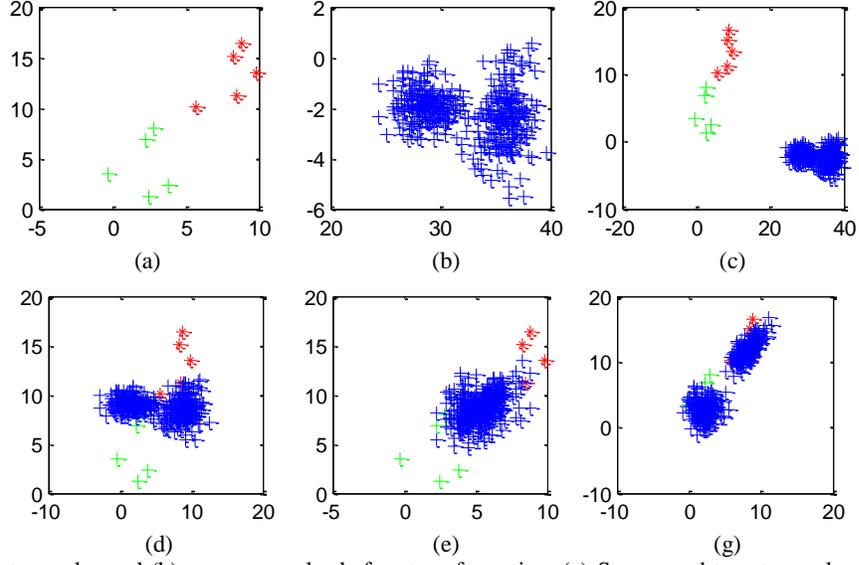

Figure 2- (a) Target samples and (b) source samples before transformation. (c) Source and target samples in one plot where the distributions of the both target and source domains are significantly different. Source and target samples are transformed to a new domain such that they have (d) minimum distance between means of both domain (d) same marginal distribution (g) same conditional distribution.

### 3.1 Domain adaptation term

As regards, the target data and source data might have been drawn from different distributions. Hence, it is necessary to find a new space in which the target and source data have a similar distribution. Up to know, some parametric and nonparametric criteria have been used to measure the distance between data distributions. In the previous approaches, this measure is defined such that the marginal distribution to be close. In this work, we propose a new criterion which measures the difference between the conditional distributions over *x* of the target and source samples. One of the measures to compute the difference of data distributions is a Maximum Mean Discrepancy (MMD) [29] criterion. In MMD, the distance between the means of samples from the two domains in the Reproducing Kernel Hilbert Space (RKHS) is calculated. In our approach, to define a new criterion which measures the difference between the conditional distributions, MMD criterion is extended to incorporate the class label of the target and source samples. Therefore, in this work, the domain adaptation term is defined as follows:

$$\hat{d}(\Im(T),\Im(S)) = \left\| \frac{1}{N_t^+} \sum_{i=1}^{N_t} y_i^t \varphi(x_i^t) - \frac{1}{N_s^+} \sum_{i=1}^{N_s} \hat{y}_i^s \varphi(x_i^s) \right\|^2 + \\ \left\| \frac{1}{N_t^-} \sum_{i=1}^{N_t} (1-y_i^t)\varphi(x_i^t) - \frac{1}{N_s^-} \sum_{i=1}^{N_s} (1-\hat{y}_i^s)\varphi(x_i^s) \right\|^2 \quad (6)$$

where $y^t$ and $\hat{y}^s$ respectively denote the true label of target samples and the hidden label of source samples. It should be noted that the source samples are unlabeled. Also, $N_t^+$ ($N_t^-$) represents the number of positive (negative) target samples and $N_s^+$ ($N_s^-$) represents the number of positive (negative) source samples. In our approach, the transformation function $\Im(.)$ is defined by $\varphi(.)$ which maps the input vector $x$ into a point in the transformed feature space. Based on the derivation of Duan et al. [13], Equation 6 can be written as follows:

$$\hat{d}(\Im(T), \Im(S)) = \operatorname{tr}(\phi'\phi I\, y'y\, I) + \operatorname{tr}(\phi'\phi I\, (1-y)'(1-y)\, I) \tag{7}$$

Where we have $y = \{\{y_i^t\}_{i=1}^{N_t}, \{\hat{y}_i^s\}_{i=1}^{N_s}\}$. In this work, the transpose of matrix is denoted by the superscript '. Let $S^+ = s^+(s^+)'$ and $S^- = s^-(s^-)'$, where $s^+$ is a column vector with $N_t + N_r$ entries that is defined as follows:

$$s^+ = Iy' \tag{8}$$

where $I$ is a diagonal matrix in which the first $N_t$ elements are set to $\dfrac{1}{N_t^+}$ and the remaining elements are set to $-\dfrac{1}{N_s^+}$. In a similar way, $s^-$ is a column vector with $N_t + N_s$ entries which is defined as follows:

$$s^- = I(1-y)' \tag{9}$$

Therefore, by substituting Equations 8 and 9 in Equation 7 and the kernel function definition $K = \phi'\phi$, we have:

$$\hat{d}(\Im(T), \Im(S)) = \operatorname{tr}(KS^+) + \operatorname{tr}(KS^-) \tag{10}$$

This reformulation is important, because it introduces the kernel function in the domain adaptation term.

### 3.2 Model risk function

In this section, the model's risk on the source and target samples is defined. In our approach, we use the structural risk functional of SVM to define $\hat{R}_T(h_\Im) + \hat{R}_S(h_\Im)$. Hence, we have:

$$\begin{aligned}\hat{R}_T(h_\Im) + \hat{R}_S(h_\Im) &= \alpha'\vec{1} - \frac{1}{2}(\alpha \circ (2y-1))'K(\alpha \circ (2y-1)) \\ \text{s.t.} \quad & 0 \leq \alpha \leq C\vec{1} \\ & \alpha'(2y-1) = 0\end{aligned} \tag{11}$$

where the hypothesis $h$ is parameterized by $\alpha$ and $\alpha = [\alpha_1\ \alpha_2\ \cdots\ \alpha_{N_t}\ \alpha_{N_t+1}\ \ldots\ \alpha_{N_t+N_s}]$ is a vector of dual variables which is assigned to each sample. Since, the labels of the samples in SVM optimization function should be -1 or 1 and $y \in \{0,1\}^{N_t+N_s}$, $(2y-1)$ in Equation 11 is used. Also, $C$ is the regularization parameter. One of the main contributions of this paper is that the source data as well as the target data in the transformed space are incorporated to learn the best classifier of the target task.

Thus, finding the optimal $h$ and $\Im(.)$ in Equation 1 is equivalent to finding the vector $y$, vector $\alpha$ and the kernel matrix $K$ that optimizes the following objective function:

$$\min_{y} \min_{K} \max_{\alpha} \{ \mathrm{tr}(KS^+) + \mathrm{tr}(KS^-) \\ + \theta(\alpha'1 - \frac{1}{2}(\alpha \circ (2y-1))'K(\alpha \circ (2y-1))) + \lambda \|yl - \tilde{y}\|^2 \} \quad (12)$$

Since, $y$ is considered as a variable, this term $\lambda\|yl - \tilde{y}\|^2$ is added to Equation 12 where it encourages the estimated label of the target data to be similar to the true target label. In other words, this term incorporates the knowledge of the true label of target data in the proposed energy function. Parameter $\lambda$ is a control parameter which balances the importance of the term. Also, $\tilde{y}$ is defined as follows:

$$\tilde{y} = \begin{bmatrix} y^t \\ 0_{N_s \times 1} \end{bmatrix} \quad (13)$$

where $0_{N_s \times 1}$ is a zero vector with $N_s \times 1$ dimension.

## 4 OPTIMIZATION

In this section, at first, the convexity of the proposed objective function is investigated and then the optimization algorithm is introduced. In this paper, instead of using semi-definite programming to learn kernel matrix $K$, similar to [13], it is assumed that the kernel function $K$ is a linear combination of a set of base kernel functions $\sum_{m=1}^{M} d_m k_m$ where $k_m$ is the $m^{\text{th}}$ base kernel matrix, $M$ is the number of base kernels and $\{d_m\}_{m=1}^{M}$ are the coefficients of the base kernels which are unknown. Hence, Equation 12 can be rewritten as follows:

$$\min_{y} \min_{d \in \Re^M} \max_{\alpha} L(y,d,\alpha) = \min_{y} \min_{d \in \Re^M} \max_{\alpha} \{ tr((\sum_{m=1}^{M} d_m k_m) Iyy'I) + tr((\sum_{m=1}^{M} d_m k_m) I(1-y)(1-y)'I) +$$
$$\theta(\alpha'1 - \frac{1}{2}(\alpha \circ (2y-1))'(\sum_{m=1}^{M} d_m k_m)(\alpha \circ (2y-1))) + \lambda \|yl - \tilde{y}\|^2 \} \tag{14}$$
s.t. $0 \leq \alpha \leq C$
$\alpha'(2y-1) = 0$
$d\vec{1} = 1$

This optimization function leads to a non-convex optimization problem. To present an optimization algorithm we rewrite it first. To do this, Equation 14 is rewritten as follows:

$$\min_{y} \min_{d \in \Re^M} \max_{\alpha} L(y,d,\alpha) = \min_{y} L(y) = \min_{y} h(y) + \lambda \|yl - \tilde{y}\|^2 \tag{15}$$

where $h(y)$ is defined as:

$$h(y) = \min_{d \in \Re^M | d'1=1} \max_{\alpha} \{ tr((\sum_{m=1}^{M} d_m k_m) Iyy'I) + tr((\sum_{m=1}^{M} d_m k_m) I(1-y)(1-y)'I) +$$
$$\theta(\alpha'1 - \frac{1}{2}(\alpha \circ (2y-1))'(\sum_{m=1}^{M} d_m k_m)(\alpha \circ (2y-1))) \} \tag{16}$$
s.t. $0 \leq \alpha \leq C$
$\alpha'(2y-1) = 0$

By letting $p_m^+ = tr(k_m S^+)$ and $p^+ = [p_1^+, p_2^+, ..., p_M^+]'$ and in a similar way $p_m^- = tr(k_m S^-)$ and $p^- = [p_1^-, p_2^-, ..., p_M^-]'$, we have:

$$h(y) = \min_{d} \max_{\alpha} \{ \frac{1}{2}(d'p^+) + \frac{1}{2}(d'p^-) + \theta(J(d,y)) \} \tag{17}$$
s.t. $0 \leq \alpha \leq C$
$\alpha'(2y-1) = 0$

where $J(d,y)$ is defined as follows:

$$J(d,y) = \alpha'1 - \frac{1}{2}(\alpha \circ (2y-1))'(\sum_{m=1}^{M} d_m k_m)(\alpha \circ (2y-1)) \tag{18}$$
s.t. $0 \leq \alpha \leq C$
$\alpha'(2y-1) = 0$

In [13], it is shown that Equation 16 is convex. As a result, once the label information is specified for the source data, Equation 15 becomes convex. In practice, we minimize Equation 15 using a coordinate descent approach. To do this, the following two steps are repeated until convergence is reached:

1) Optimize *d* and *α*: In this step, *y* is assumed to be fixed and $L(y,d,\alpha)$ is optimized over *d* and *α* by solving the convex optimization problem defined by $h(y)$. To do this optimization, the introduced procedure in Algorithm 1 (lines 14-17) is applied [13].

2) Obtain *y*: optimize $L(y,d,\alpha)$ over *y* by fixing *d* and *α*. To do this step, it is optimized using CVX package which is demonstrated in the following (Algorithm 1- line 19).

To do the optimization of $L(y,d,\alpha)$ over *d* and *α* in step 1, at first, the vector *α* of SVM objective is obtained:

$$\alpha^* = \arg\max_{\alpha} \{\theta(\alpha'1 - \frac{1}{2}(\alpha \circ (2y-1))'(\sum_{m=1}^{M} d_m k_m)(\alpha \circ (2y-1)))\} \quad (19)$$

$$s.t. \quad 0 \leq \alpha \leq C$$
$$\alpha'(2y-1) = 0$$

Then, similar to [13], SVM parameters are fixed and kernel parameters (*d*) is obtained. To do this, the iterative algorithm of the reduced gradient method suggested in [30], is used to update *d* which is given in the following:

$$d_{t+1} = d_t - \eta_t g_t \quad (20)$$

where $g_t = (\nabla^2 h)^{-1} \nabla h$ in which the gradient vector and the Hessian matrix are defined as $\nabla h = (p^+(p^+)' + \varepsilon I)d + (p^-(p^-)' + \varepsilon I)d + \nabla J(d,y)$ and $\nabla^2 h = (p^+(p^+)' + \varepsilon I) + (p^-(p^-)' + \varepsilon I)$, respectively. These two steps are repeatedly done until convergence is reached. Here, when the base kernels coefficients do not change between the consecutive steps, convergence happens.

In step 2, to obtain *y*, $L(y) = h(y) + \lambda \|yl - \tilde{y}\|^2$ should be optimized. Since $L(y)$ is convex, we use a standard optimization package called CVX [31] to solve this problem. The process of optimization between step 1 and step 2 is repeated until convergence is reached.

After convergence, we have a relatively strong local optimum in the sense that step 1 searches over an exponentially-large space of latent values of parameters *d* and *α* while step 2 searches over an exponentially-large space of latent class labels of source data.

| Algorithm 1- STSVM Algorithm |
|---|
| 1: **Inputs:** Target data $T = \{x_i^t, y_i^t\}_{i=1}^{N_t}$ where $x^t \in \chi$ and $y^t \in \{0,1\}$ |
| 2:         Source data $S = \{x_i^s\}_{i=1}^{N_s}$ |
| 3: Output: $f: \chi \to \{0,1\}$ using target and source data |
| 4: **Begin** |
| 5:      For i=1:*M* |
| 6:         Compute $k_i$ ($i^{th}$ base kernel) |
| 7:      End |
| 8:      Initialize $d = \frac{1}{M}\vec{1}$ and compute kernel matrix $K$ |
| 9:      Learn a SVM model on target data |
| 10:     Initialize $y = \begin{bmatrix} y^t \\ \hat{y}^s \end{bmatrix}$ where $\hat{y}^s$ is obtained by applying the learned model of SVM on source data |
| 11:     Compute $S^+$ and $S^-$ using Equations 8 and 9 |
| 12:     **Loop** until convergence |
| 13:        // update $d$ and $\alpha$ by fixing $y$ |
| 14:        **Loop** until convergence |
| 15:           Solve $\alpha$ of SVM objective in Equation 19 |
| 16:           Update $d$, weights of base kernels, using 20 |
| 17:        **End** |
| 18:        // update $y$ by fixing $d$ and $\alpha$ |
| 19:        optimize $h(y)$ over $y$ using CVX package |
| 20:     **End** |
| 21: **End** |

## 5 EXPERIMENTS

In this section, we carried out several experiments on different data sets. We analyzed two sets of experiments: 1) a real problem in the framework of image classification where the target data is chosen from the MSRC dataset and the source data are sampled from the LMO dataset, and 2) a real problem in object recognition where the target and source samples are drawn from Caltech-256 dataset. In this experiment, the source domain knowledge can be related or unrelated to the target domain.

To show the effectiveness of the proposed method, we compared some variants of our proposed method with the supervised learning method SVM in which the kernel function is considered as the sum of 16 base kernels. We also compared our method with DTSVM [13], where it is similar to our approach in using the domain adaptation term and the model risk function but it differs from our approach in that it uses the standard MMD measure while in our approach, a new MMD measure is defined such that it incorporates the class label of target and source samples.

Also, in their formulation, in the model risk function, the source samples are not incorporated. We also compared our approach with the state-of-the-art domain adaptation approaches TCA [22] and CODA [15]. In TCA, without losing any generality, to have a fair comparison, the labeled source and target samples are considered as the source domain and the unlabeled test samples are considered as the target domain. Also, in CODA, the source samples and target samples are considered as labeled samples. CODA is an iterative algorithm in which in each iteration, *c* most confident predictions on unlabeled test samples are moved to the labeled samples for the next iteration. In this paper, two versions of STSVM are used: STSVM in which iterative approach for source label updating is removed which is called STSVM-I, and full STSVM. In other words, in STSVM-I, in Algorithm 1, line 19 is removed. It should be noted that, in our experiments, the source code of 'CODA' and TCA are downloaded[1] and used. But, DTSVM is implemented by ourselves.

## 5.1 Experimental setting

In STSVM and all the compared kernel based methods, four types of base kernels are used:

1) Gaussian kernel ( $k(x_i, x_j) = \exp(-\gamma \| x_i - x_j \|^2)$ ),

2) Laplacian kernel ( $k(x_i, x_j) = \exp(-\sqrt{\gamma} \| x_i - x_j \|)$ ),

3) Inverse square distance kernel ( $k(x_i, x_j) = 1/(\gamma \| x_i - x_j \|^2 + 1)$ ), and

4) Inverse distance kernel ( $k(x_i, x_j) = 1/(\sqrt{\gamma} \| x_i - x_j \| + 1)$ ).

The kernel parameter for Gaussian kernel is set to four different values which are obtained as $(1.2^\sigma)/d$ where $\sigma$ is set to {2, 2.5, 3, 3.5} and for the other three types, $\sigma$ is set to {3, 3.5, 4, 4.5}. Therefore, we have 16 base kernels in total (*M*=16). Moreover, in this paper, in all experiments, the parameter $\varepsilon$ is set to $10^{-4}$, SVM regularization parameter (*C*) is set to 10 and $\theta$ is set to 1. In SVM, the coefficients of base kernels are set to 1/16.

## 5.2 Image classification

In this section, we evaluate our approach on image classification datasets. To do this, we utilize the two following datasets:

---

[1] http://www.cse.wustl.edu/~mchen/code/coda.tar
  http://www.cse.ust.hk/TL/code/TCA-AAAI2012-version2.0.zip

1) MSRC dataset [32] which has 512 images which are divided into train and test sets based on the standard split method. It contains 20 image categories with up to 14 images in each category of train set and 16 images in each category of test set. Image categories of MSRC dataset are grass, tree, building, airplane, cow, face, car, bicycle, sheep, flower, sign, bird, book, chair, cat, dog, street, boat, human and water. Some images of the MSRC dataset are shown in Figure 2-a.

2) LabelMe Outdoor (LMO) dataset [33] has 2688 outdoor images. It contains 8 image categories which include coast, forest, highway, inside city, mountain, open country, street and tall building. Some images of the LMO dataset are shown in Figure 2-b.

To describe each image, Spatial Pyramid Matching [34] is used. To do this, key points are densely sampled with the sampling step 8 pixels in horizontal and vertical directions. To describe each key point SIFT [35], Hue descriptor [36] and 17-filter bank are used. For each descriptor, a 200 visual word dictionary is constructed independently.

In this experiment, for each image category in MSRC dataset, the related images from the training set of MSRC dataset are chosen as positive images and 10 samples from the remaining of training images are randomly selected as negative target samples. Also, we randomly select 1000 unlabeled images from the LMO as the source data set. Then, for each object category, the proposed approach is trained on the source and target dataset. In the test step, the learned model is applied to test set. Each experiment is repeated 10 times and then the mean of accuracy is reported which is shown in Table 1. In Table 1, the mean of classification accuracy and its standard deviation of each image category are presented. Also, our approach is compared to SVM, CODA and TCA and STSVM_I. As it is shown in Table 1, our approach receives 72.52% average accuracy with a standard deviation 7.39%. The proposed methods are consistently better than other compared methods, which demonstrate their effectiveness in the task of transferring knowledge in image classification. Second, STSVM_I method does not have a satisfactory classification performance. This is because it is assumed that the source and target data come from a different distribution. For initial labeling of source data in our approach, at first, we learn a SVM classifier with the target data and then the learned SVM model is applied on the source data. Hence, the initial labeling of the source data is not reliable. Consequently, it effects on the final results. It should be noted that although LMO images and MSRC images are from different distributions (see Figure 2), our proposed approach can improve the classification accuracy.

In this experiment, for each image category, the test set contains 16 positive images and 304 negative images. The standard technique used to assess the performance of the model is inappropriate for the unbalanced test data.

Therefore, to have a fair comparison, the geometric mean is also reported. In Table 2, the average of geometric mean and its standard deviation of each image category are presented. As it is shown, our approach, in many categories, receives higher accuracy than the other approaches. The main reason is for this that, in our approach, the domain adaptation term and the model risk function simultaneously consider the discriminative information contained in the source data. It should be noted that in CODA, the source and target samples are all labeled; nevertheless, our approach in many categories receives better accuracy. As it is shown in Table 2, our approach, in structured class label, has received more improvement in performance with respect to the other categories. It should be noticed that many structured class label such as sheep, cat, human, sign, chair and bird do not exist in the source domain; nevertheless, knowledge transfer is done in a good way with respect to the other categories. In our approach, the image-level label contains a single dominant class label. However, an image might contain more than one dominant class label (see Figure 4). The incompleteness of image-level label for an image means that this image may contain the other class labels which cannot be assigned to it. This incompleteness affects the performance of the other approaches. Image level labels of images in Figure 4-a and Figure 4-b are 'grass' and 'cow' respectively. However, as it is shown, there are high similarities between these images. The same situation is also established for Figures 4-c and Figures 4-d. It leads to a decrease in performance in some categories like 'boat' and 'grass'.

Also, to study the effectiveness of the number of positive samples in the target data, we design an experiment. In this experiment, positive target data are sampled from the semantic class 'cow'. Figure 5 shows the true positive rate of each method as the number of positive samples ranging from 1 to 14. We can see that STSVM can classify the large number of positive samples correctly. When the number of positive samples is 1, SVM cannot classify any positive test samples; however, our approach receives 23% average true positive rate.

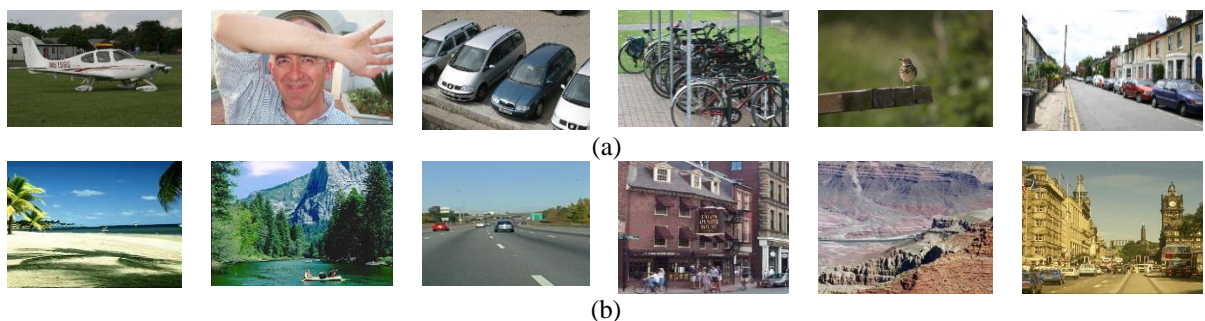

Figure 3- (a) sample images of the MSRC dataset and (b) sample images of the LMO dataset.

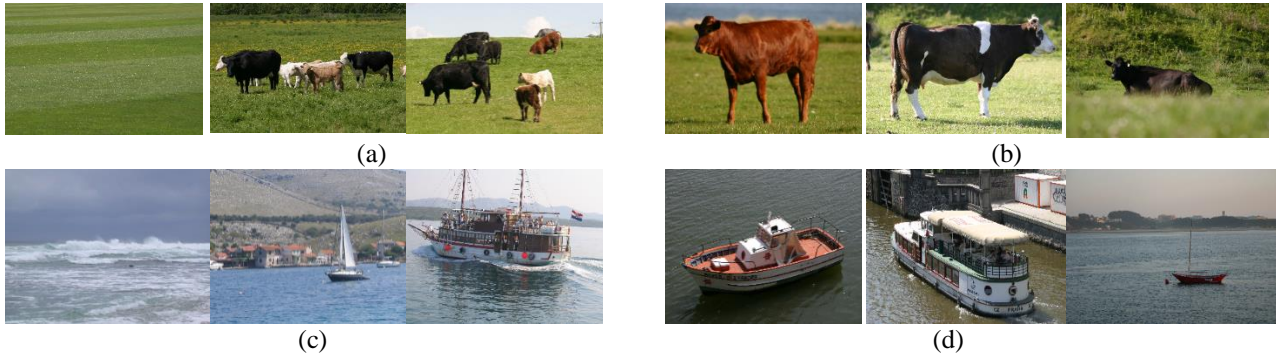

Figure 4- Sample images of the MSRC dataset with image level label of (a) 'grass' (b) 'cow' (c) 'water' and (d) 'boat'. Due to the incompleteness of image-level, there are high similarities between 'cow and grass' and 'water and boat' images.

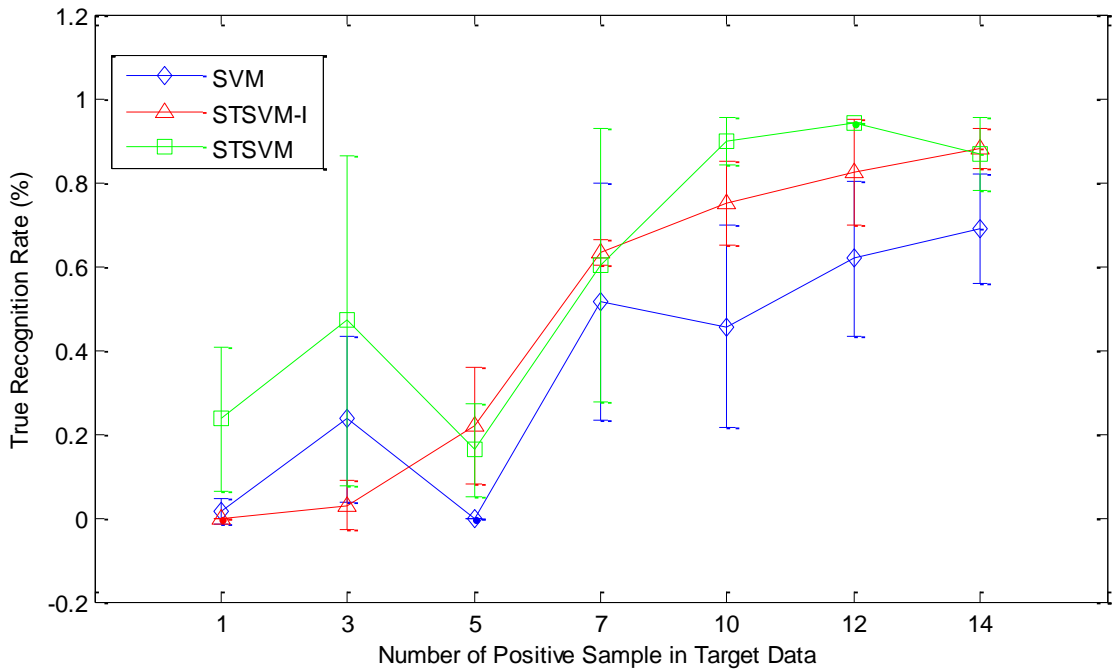

Figure 5- Performance comparison of true recognition rate with respect to the number of positive samples in target data. Positive target data are sampled from the semantic class 'cow' MSRC dataset and 1000 samples as source data are randomly drawn from LMO dataset.

Table 1- Average classification accuracy of our proposed method and some other methods when the target data are drawn from MSRC dataset and source data are sampled from LMO dataset (all numbers are given in percentage).

| Method<br>Class | SVM | CODA | DTSVM | TCA | STSVM_I | STSVM |
|---|---|---|---|---|---|---|
| Grass | **78.34±1.86** | 51.29±8.97 | 76.48±3.92 | 61.98±18.12 | 45.50±4.60 | 72.24±6.96 |
| Tree | **85.71±3.47** | 67.84±9.80 | 82.64±4.70 | 58.34±13.05 | 48.71±4.52 | 78.62±8.84 |
| Building | 52.59±5.26 | 53.46±10.47 | 50.41±6.82 | **78.48±8.81** | 44.06±14.32 | 64.66±9.65 |
| Airplane | 81.31±4.72 | 77.88±6.08 | **89.18±4.32** | 58.34±8.77 | 54.96±4.32 | 84.69±2.82 |
| Cow | 62.36±4.33 | 53.69±8.42 | 62.44±6.33 | **72.92±15.12** | 39.19±5.94 | 68.15±8.06 |
| Face | 78.24±4.64 | 71.51±4.55 | **82.22±7.49** | 74.74±8.88 | 51.57±10.17 | 79.68±6.68 |
| Car | 81.87±6.41 | 61.25±9.50 | 80.70±5.84 | 77.48±10.19 | 49.31±5.61 | **82.71±7.13** |
| Bicycle | 78.27±6.71 | 51.15±5.96 | 76.03±4.45 | 80.69±7.48 | 50.09±5.86 | **81.94±6.76** |
| Sheep | 72.70±5.33 | 54.61±7.91 | **73.52±6.62** | 68.36±15.11 | 44.94±4.39 | 69.31±9.55 |
| Flower | **89.10±2.87** | 56.09±12.23 | 80.67±5.85 | 86.06±4.53 | 35.73±5.33 | 75.24±8.32 |
| Sign | 50.97±8.54 | 51.80±8.20 | 49.87±9.51 | 47.45±16.87 | 25.19±6.44 | **55.91±9.28** |
| Bird | 56.01±6.83 | 64.92±3.94 | 58.10±5.54 | 49.33±9.42 | 42.89±11.47 | **68.92±5.72** |
| Book | **92.63±1.69** | 68.47±8.64 | 90.44±5.35 | 66.54±23.59 | 63.95±10.46 | 85.82±8.11 |
| Chair | 30.83±3.76 | 48.37±7.10 | 34.89±5.42 | 45.72±17.41 | 27.02±6.37 | **50.48±7.75** |
| Cat | 64.02±7.27 | 54.21±7.27 | 65.18±6.36 | 75.21±5.06 | 39.12±15.67 | **76.54±6.61** |
| Dog | 50.58±4.82 | 50.44±6.89 | 49.97±5.70 | **76.56±9.88** | 41.94±4.33 | 67.55±6.90 |
| Street | **80.74±4.29** | 57.74±7.37 | 79.30±4.63 | 69.27±9.86 | 54.50±5.91 | 75.59±5.27 |
| Boat | 69.66±3.19 | 65.34±6.46 | 68.70±5.48 | **75.74±19.83** | 53.05±10.18 | 66.14±3.51 |
| Human | 35.77±3.28 | 49.68±5.27 | 37.94±3.61 | **68.36±10.19** | 40.63±2.50 | 56.76±16.85 |
| Water | 87.83±3.66 | 79.37±4.64 | **92.16±2.58** | 72.17±16.77 | 75.59±5.12 | 89.49±3.05 |
| Mean | 68.97±4.64 | 59.46±7.48 | 69.04±5.53 | 68.19±12.45 | 46.39±7.17 | **72.52±7.39** |

Table 2- Average geometric mean (Gmean) of our proposed method and some other methods when the target data are drawn from MSRC dataset and source data are sampled from LMO dataset (all numbers are given in percentage).

| Method Class | SVM | CODA | DTSVM | TCA | STSVM_I | STSVM |
|---|---|---|---|---|---|---|
| Grass | 83.54±2.88 | 73.99±3.79 | **83.50±2.89** | 80.46±3.69 | 78.86±5.55 | 82.65±3.66 |
| Tree | **90.34±3.15** | 85.67±11.09 | 90.09±2.99 | 53.49±31.71 | 86.03±3.31 | 90.01±1.69 |
| Building | 67.69±3.85 | 67.59±5.12 | 67.08±4.57 | 11.52±17.93 | 62.30±6.83 | **72.8±4.42** |
| Airplane | 94.33±2.37 | 89.70±12.18 | 94.10±2.45 | 71.75±40.11 | **95.68±2.81** | 94.22±1.68 |
| Cow | 61.24±4.33 | 71.91±6.53 | **76.05±4.44** | 69.63±9.72 | 55.46±7.02 | 75.25±7.34 |
| Face | **88.98±4.00** | 77.21±4.28 | 88.69±4.28 | 35.04±31.62 | 86.35±4.25 | 86.98±3.99 |
| Car | **89.70±3.43** | 79.99±17.66 | 89.19±3.45 | 25.63±38.09 | 85.43±7.72 | 86.18±2.82 |
| Bicycle | 86.99±2.66 | **88.94±10.83** | 86.42±2.72 | 66.36±37.81 | 88.45±7.83 | 88.70±1.89 |
| Sheep | 84.8±4.26 | 80.05±12.31 | 84.82±4.28 | 82.84±2.23 | 86.01±2.36 | **89.04±1.61** |
| Flower | **93.32±1.64** | 79.98±12.76 | 93.14±1.63 | 78.49±9.87 | 90.14±3.33 | 87.46±3.41 |
| Sign | 66.11±7.48 | 65.93±13.42 | 65.73±8.27 | 26.48±7.59 | 56.61±6.28 | **73.76±4.39** |
| Bird | 69.71±4.54 | 72.74±5.81 | 70.20±4.31 | 58.67±1.47 | 69.82±6.05 | **80.95±3.22** |
| Book | **95.12±2.36** | 90.82±8.05 | 94.79±3.49 | 66.65±38.55 | 87.83±14.51 | 92.74±2.48 |
| Chair | 55.34±4.58 | 72.22±13.20 | 55.36±4.65 | 40.99±14.90 | 54.57±11.21 | **73.55±4.14** |
| Cat | 78.28±3.79 | 82.86±16.17 | 78.51±4.10 | 15.64±23.15 | 76.38±5.66 | **83.78±1.89** |
| Dog | 63.17±3.17 | **77.64±15.58** | 63.06±3.24 | 26.51±24.85 | 61.30±5.52 | 71.83±3.28 |
| Street | **77.51±5.93** | 69.21±3.61 | 76.99±6.06 | 32.29±23.00 | 69.95±5.20 | 76.58±3.76 |
| Boat | **75.36±3.38** | 69.71±11.41 | 75.22±3.63 | 15.18±22.40 | 72.88±2.31 | 72.92±3.64 |
| Human | 55.80±3.72 | 65.39±17.39 | 55.16±3.46 | 28.55±18.45 | 47.35±7.35 | **66.92±5.75** |
| Water | **91.63±2.57** | 80.63±8.23 | 91.53±2.69 | 9.70±21.68 | 81.99±4.64 | 82.75±5.10 |
| Mean | 78.45±3.70 | 77.11±10.47 | 78.98±3.88 | 44.79±20.94 | 74.67±5.99 | **81.45±3.51** |

To investigate the role of kernels in kernel based methods, we design a new experiment. In this experiment, the number of base kernels is set to {4 (four Gaussian base kernels), 8 (four Gaussian and four Laplacian base kernels), 12 (four Gaussian, four Laplacian and four Inverse square distance base kernels), and 16 (four Gaussian, four Laplacian, four Inverse square distance and four Inverse distance base kernels)}. We use accuracy and geometric mean to evaluate the performance for each set of kernels. The obtained results are shown in Figure 6. As it is shown, our proposed approach, even with the four base kernels receives a more reasonable performance with respect to the other approaches.

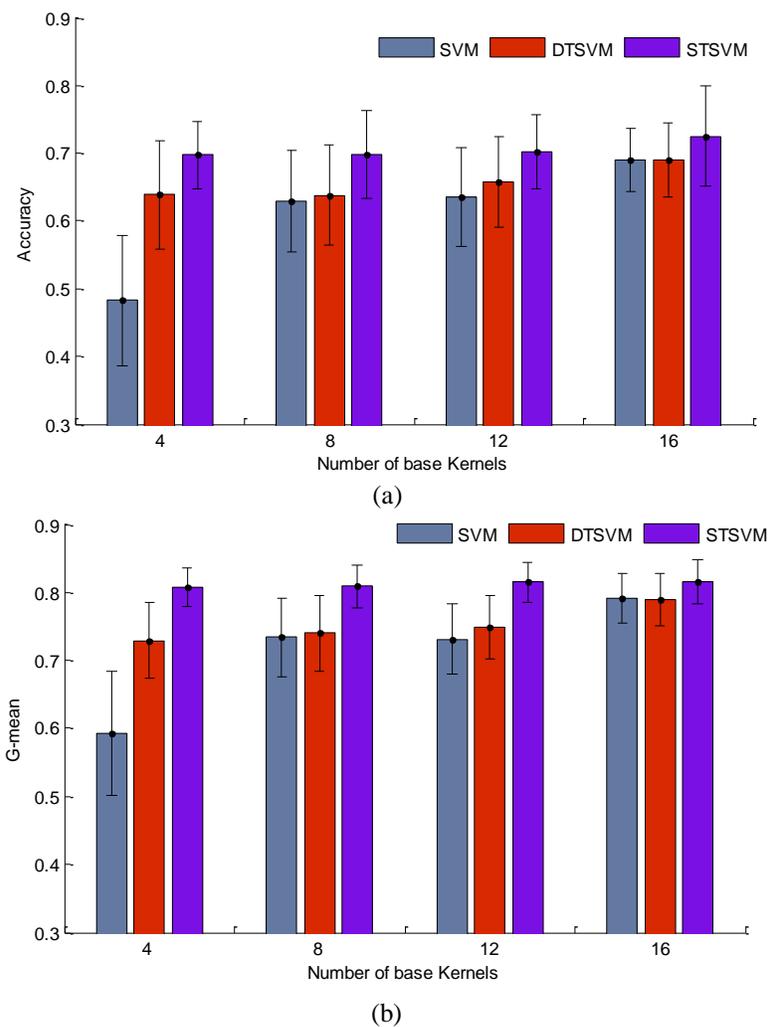

Figure 6- (a) Recognition accuracy and (b) geometric mean of SVM, DTSVM and STSVM on MSRC+LMO dataset under different number of base kernels.

In this paper, to study the time complexity of our approach, we measure the running time of our approach and the baseline methods on the same input and the same machine for both training and test phases. All algorithms are run on the MSRC+LMO dataset and then results are averaged. The obtained results are given in Table 3. As it is shown, our approach does better than TCA and worse than CODA.

Table 3- Average of run time of our approach and the baseline approaches (time is in second).

| Methods | SVM | DTSVM | TCA | CODA | STSVM |
|---|---|---|---|---|---|
| Runtime | 0.0007 | 19.98 | 849.64 | 286.76 | 393.78 |

## 5.3 Object recognition

In this section, experiments are run on the subsets of the Caltech-256 database [37]. Caltech-256 dataset is a set of 256 object categories containing a total of 30607 images. In each object category, the minimum number of images is between 31 to 80. Also, it contains a background set which consists of random images. In this paper, similar to the setting of [8], different groups of related and unrelated categories are taken. To consider the related classes, two sets of 6 classes are intended: set 1 that contains 'bulldozer', 'firetruck', 'motorbikes', 'schoolbus', 'snowmobile' and 'car-side' classes and set 2 includes 'cake', 'hamburger', 'hot-dog', 'ice-cream-cone', 'spaghetti' and 'sushi' classes. The unrelated classes contain 'dog', 'horse', 'zebra', 'helicopter', 'fighter-jet', 'motorbikes', 'car-side', 'dolphin', 'goose' and 'cactus' classes which is called set 3.

In this paper, to describe each image, the pre-computed features of [10] are used which are available on the web. In our approach, similar to [8], four different image descriptors: PHOG Shape Descriptors [38], Appearance Descriptors [39], Region Covariance [40] and Local Binary Patterns [41] are used. All of the image descriptors were computed in a spatial pyramid which we considered as the first level.

In this section, we design two different experiments. In the first experiment, we pair any two classes and then the target data is provided from one class and the source data is provided from the other class. In this case, 10 samples from the target class are randomly chosen as the positive target samples and 50 images from the background dataset are chosen as negative target samples. Also, 50 samples from the source class are randomly chosen as the source data. The test set consists of 100 images, half from the background and half from the target class. In this case, each experiment is repeated 10 times and then the mean of the accuracy is reported. In Figure 7, the classification accuracy of each class pair of set 1 (related classes) is presented. Also, the classification accuracy of each class pair of set 2 (related classes) is shown in Figure 8. These experiments show how the knowledge is transferred between the two classes. As it is shown, STSVM, in many pairs, does better than STSVM-I, DTSVM and SVM.

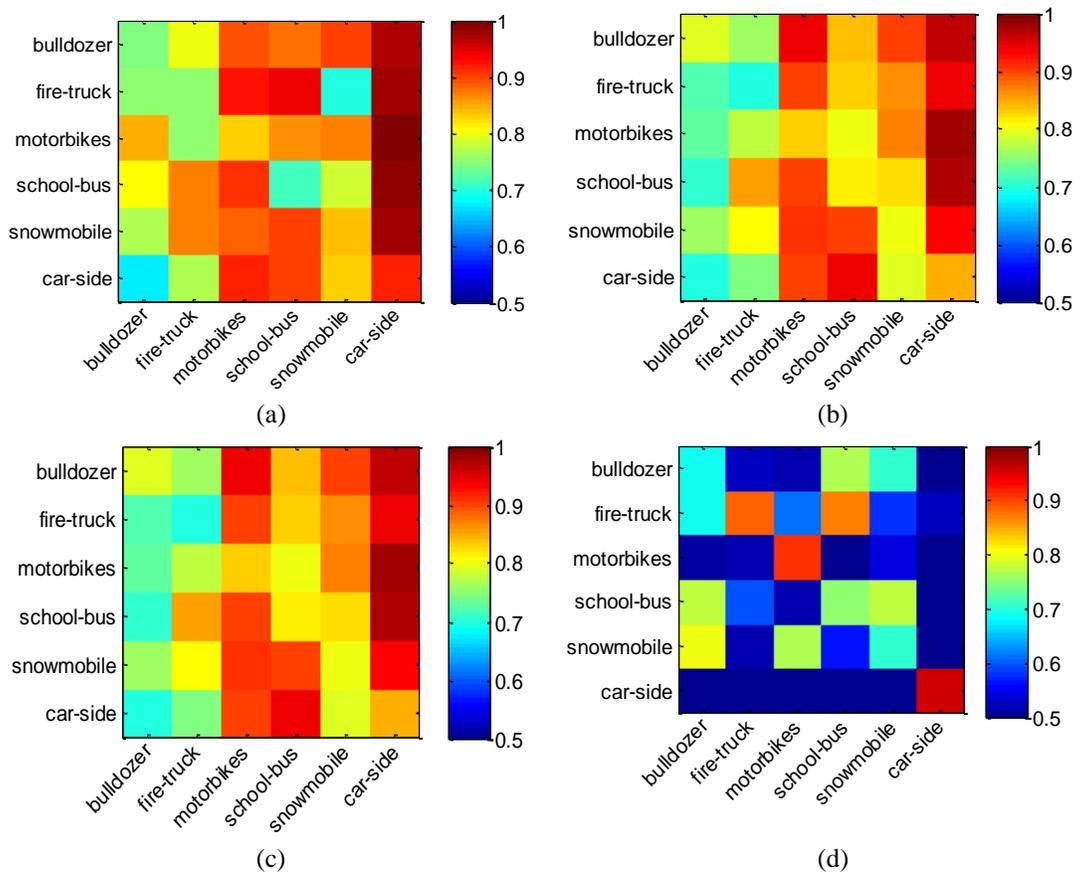

Figure 7- classification accuracy of each class pair of set 1 by (a) STSVM (b) STSVM-I (c) DTSVM (d) SVM. Our approach in many pair classes of set 1 (one as a target and the other as source class) achieves the highest recognition rate. Each experiment is repeated 10 times and then the mean of the recognition rates is reported.

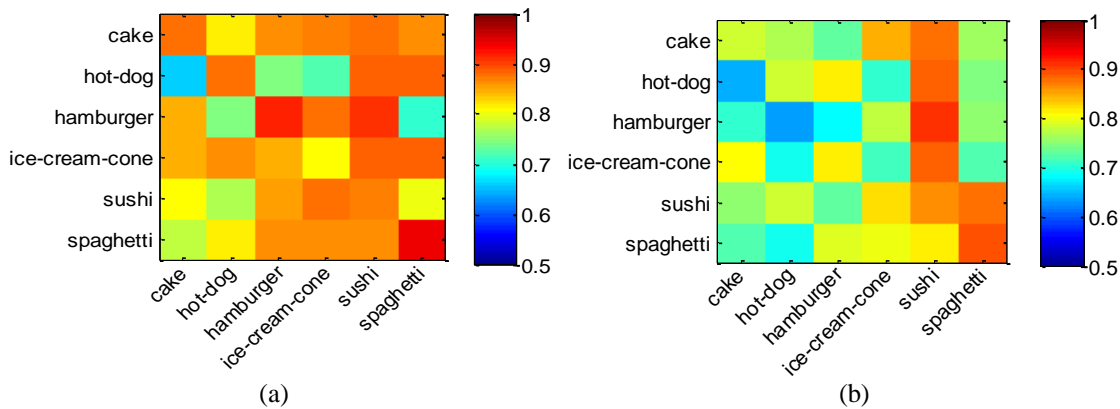

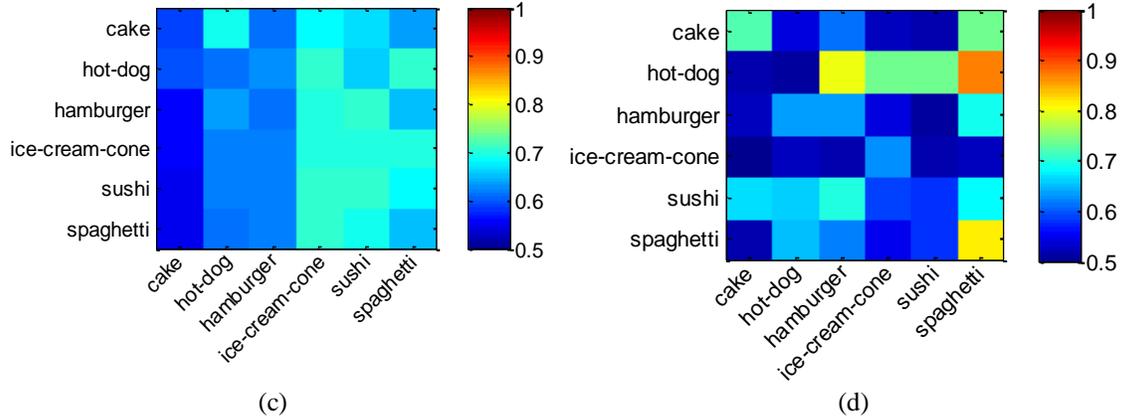

(c)                                            (d)

Figure 8- classification accuracy of each class pair of set 2 by (a) STSVM (b) STSVM-I (c) DTSVM (d) SVM. Our approach in many pair classes of set 2 (one as a target and the other as source class) achieves the highest recognition rate. Each experiment is repeated 10 times and then the mean of the recognition rates are reported.

Also, all pairs in which target class is the same are considered and their mean accuracy and its standard deviation are reported in Table 4. The obtained results are compared with DTSVM, CODA and baseline SVM. Our approach in all classes of set 1 except 'school-bus' and 'car-side' improves the classification accuracy. In 'school-bus' and 'car-side', DTSVM receives a better accuracy with respect to our approach about 2.02% and 0.95%, respectively. The most improvement in accuracy is obtained in 'firetruck' class which is about 13%. In set 2, our approach, in all classes, improves the performance. By comparing the performance of STSVM-I with DTSVM, it is concluded that incorporating class labels in domain adaptation and incorporation of target and source data in classifier optimization have a significant impact on performance improvement. As it is noted, in STSVM-I, the variable $y$ is set to $y = \begin{bmatrix} y^t \\ \hat{y}^s \end{bmatrix}$ where $\hat{y}^s$ is obtained by applying the learned model of SVM on source data and the updating of $y$ is ignored. By comparing the results of STSVM-I with STSVM in Table 4, the effect of updating procedure of $y$ on the performance of the proposed algorithm is shown. As it is represented in Table 4, even in the unrelated categories, the unrelated source samples can improve the classification accuracy of the target samples. The average classification accuracy of each set in our approach, DTSVM, CODA and STSVM-I approaches are shown in Figure 9. Compared with DTSVM, the average classification accuracy improvements are 11.75% and 8.58% for the related and unrelated sets respectively.

Table 4- Comparison between our STSVM algorithm, STSVM-I, DTSVM, CODA and SVM in the first experiment setting (All numbers are given in percentage).

| Set No. | Class | SVM | DTSVM | TCA | CODA | STSVM-I | STSVM |
|---|---|---|---|---|---|---|---|
| Set 1 | bulldozer | 66.17±11.93 | 72.00±2.91 | 50.70±4.48 | 72.86±4.14 | 73.57±6.89 | **76.63±4.67** |
| | firetruck | 59.33±13.75 | 67.33±4.42 | 49.10±3.84 | 71.27±5.10 | 77.45±6.08 | **80.23±5.16** |
| | motorbikes | 63.83±15.35 | 88.33±0.48 | 48.97±5.25 | 85.16±4.95 | **89.83±6.91** | 89.50±4.21 |
| | schoolbus | 66.00±14.47 | **88.67±0.48** | 50.57±3.18 | 80.67±3.87 | 85.43±5.46 | 86.65±7.60 |
| | snowmobile | 63.67±10.29 | 72.17±1.26 | 49.43±4.67 | 64.66±4.88 | **84.25±7.39** | 82.13±7.67 |
| | car-side | 57.33±19.12 | **98.33±1.03** | 50.80±3.84 | 96.47±2.06 | 94.00±4.63 | 97.38±2.54 |
| Set 2 | Cake | 57.67±8.56 | 56.83±1.96 | 50.23±3.82 | 65.09±3.98 | 73.60±8.60 | **80.42±9.00** |
| | hamburger | 58.83±8.68 | 63.17±0.69 | 47.97±4.55 | 71.42±4.97 | 72.62±6.06 | **81.53±6.09** |
| | hot-dog | 64.83±6.31 | 61.83±2.82 | 50.77±3.86 | 69.04±3.98 | 76.12±5.73 | **84.80±4.34** |
| | ice-cream-cone | 59.67±7.31 | 70.17±1.08 | 49.77±5.07 | 70.83±1.79 | 77.78±5.55 | **83.62±6.29** |
| | spaghetti | 57.50±11.01 | 69.00±2.69 | 50.33±3.88 | 71.93±4.53 | 87.45±5.40 | **88.45±6.89** |
| | sushi | 72.17±8.32 | 67.17±1.87 | 50.30±3.88 | 69.43±4.63 | 79.28±5.45 | **84.65±3.03** |
| Set 3 | Dog | 53.87±4.88 | 56.77±1.67 | 49.50±3.87 | 63.48±3.39 | 63.27±4.94 | **67.32±5.00** |
| | horse | 52.20±3.36 | 55.20±0.98 | 49.66±3.65 | 66.31±1.29 | 66.43±6.32 | **66.58±5.99** |
| | Zebra | 55.80±7.56 | 62.00±2.42 | 49.36±3.84 | 75.49±5.04 | 67.33±5.85 | **75.55±8.78** |
| | helicopter | 56.70±12.92 | 75.60±3.22 | 50.48±4.12 | 76.33±4.87 | 76.03±6.12 | **81.43±4.60** |
| | fighter-jet | 57.80±12.12 | 66.20±1.95 | 49.54±4.09 | 73.74±4.30 | 69.88±6.23 | **74.31±4.81** |
| | motorbikes | 56.00±14.68 | 88.40±0.64 | 49.14±4.67 | 87.01±4.41 | 88.47±4.21 | **91.10±2.52** |
| | car-side | 54.50±13.91 | **98.80±0.75** | 50.42±6.43 | 97.59±2.68 | 91.22±5.40 | 96.37±2.98 |
| | dolphin | 51.30±6.93 | 55.30±2.38 | 49.94±3.64 | 66.43±4.47 | 64.56±7.42 | **67.45±9.30** |
| | goose | 56.60±7.31 | 57.90±2.27 | 49.32±3.67 | 65.54±1.09 | 67.51±5.76 | **67.60±7.12** |
| | cactus | 54.20±8.47 | 55.10±2.68 | 50.72±4.43 | 66.82±2.81 | **70.97±7.09** | 69.37±8.56 |

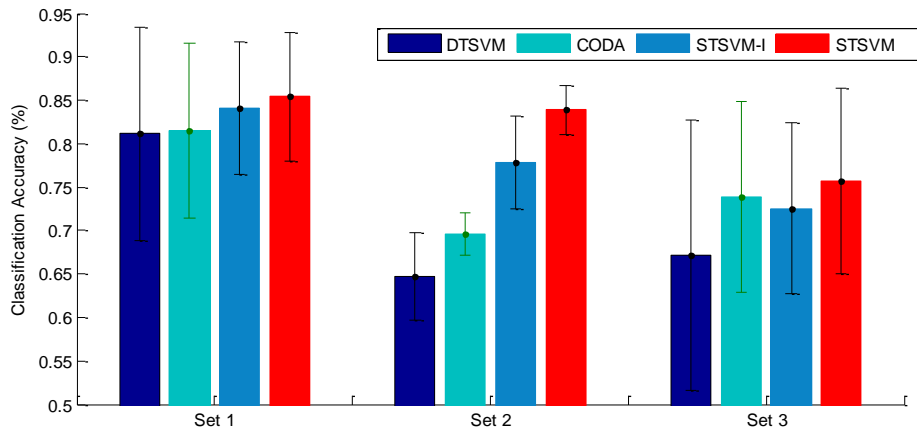

Figure 9- The average classification accuracy of each set in our approach and compared methods in the first experiment setting.

In the second experiment, in each set, one class is considered as a target domain and the other classes of the set are considered as the source domains. In this experiment, 10 samples from the target class are randomly chosen as positive target samples and 50 images from the background dataset are chosen as negative target samples. Also, 50 samples from each class of the source domains are randomly chosen as the source data. Hence, the numbers of the source data for set1, set 2 and set 3 are 250, 250 and 450 respectively. The test set consists of 100 images, half from the background and half from the target class. Each experiment is repeated 10 times and then the mean of the recognition rates is reported. In Table 5, the performance accuracy under this experiment is shown. As it is shown, our approach in all classes of set 2 and in 8 out of 10 classes of set 3 receives the better accuracy with respect to the other baseline methods. It should be noted that, in TCA and CODA, all source samples are considered as labeled samples, but they cannot receive the reasonable accuracy. The main reason is for this that the discriminative information contained in the unlabeled source data is simultaneously considered in the domain adaptation term and the model risk function.

The average classification accuracy of each set in our approach, DTSVM, CODA and STSVM-I approaches in the second experiment setting, are shown in Figure 10. In this setting, compared with CODA, the average classification accuracy improvements are 0.5% and 2.50% for the related and unrelated classes respectively. As it is shown, our approach increases performance on the unrelated source and target data more than the related one.

Table 5- Comparison between our STSVM algorithm, STSVM-I, DTSVM, CODA and SVM in the second experiment setting (all numbers are given in percentage).

| Set No. | Class | SVM | DTSVM | CODA | TCA | STSVM-I | STSVM |
|---|---|---|---|---|---|---|---|
| Set 1 | bulldozer | 64.31±3.93 | 67.60±3.75 | 70.31±3.41 | 50.40±4.04 | 53.12±5.97 | **71.03±1.26** |
| | firetruck | 51.31±6.69 | 61.09±5.84 | **73.24±5.53** | 50.00±3.46 | 51.06±7.59 | 57.39±1.03 |
| | motorbikes | 34.68±3.48 | 91.35±3.36 | 85.96±3.06 | 51.40±4.39 | 79.84±4.90 | **94.21±2.11** |
| | schoolbus | 48.27±3.21 | **89.17±3.16** | 84.14±2.60 | 50.20±6.38 | 55.34±3.43 | 87.59±0.00 |
| | snowmobile | 41.12±2.31 | 62.69±1.85 | 66.36±4.16 | 49.40±3.78 | 50.09±3.47 | **71.12±1.83** |
| | car-side | 21.03±10.87 | 97.86±1.85 | 96.85±3.15 | 48.40±3.78 | 96.23±3.16 | **98.96±0.00** |
| Set 2 | Cake | 50.43±3.75 | 58.41±3.30 | 60.23±3.71 | 50.40±1.14 | 51.36±3.87 | **64.21±2.21** |
| | hamburger | 43.91±4.03 | 61.42±4.23 | **68.09±5.50** | 50.00±3.32 | 49.68±3.46 | 66.39±1.26 |
| | hot-dog | 53.39±4.64 | 52.73±4.66 | 67.43±4.77 | 52.40±2.88 | 54.16±5.85 | **69.24±1.45** |
| | ice-cream-cone | 39.31±3.37 | 65.80±3.75 | **67.79±4.34** | 50.60±2.79 | 51.33±9.28 | 62.86±5.11 |
| | spaghetti | 47.83±5.52 | 76.30±5.12 | 70.23±2.74 | 50.40±6.99 | 50.21±5.94 | **83.68±1.05** |
| | sushi | 50.19±5.91 | **72.45±6.01** | 69.06±6.65 | 52.20±2.05 | 49.86±6.65 | 58.14±1.76 |
| Set 3 | Dog | 62.39±7.30 | 61.33±7.15 | 59.35±3.80 | 52.80±3.90 | 50.39±4.89 | **66.23±2.53** |
| | horse | 54.12±4.06 | 57.19±4.28 | 64.29±5.76 | 50.20±2.28 | 52.45±2.87 | **68.11±1.03** |
| | zebra | 52.09±6.28 | 64.61±6.29 | **75.08±5.48** | 50.40±3.51 | 52.36±7.11 | 68.91±1.33 |
| | helicopter | 31.78±5.92 | 82.24±6.14 | 75.91±4.93 | 51.20±1.30 | 53.59±4.55 | **82.32±1.90** |
| | fighter-jet | 49.05±3.74 | 68.19±4.17 | **74.23±2.91** | 47.60±6.11 | 50.25±9.76 | 72.14±2.75 |
| | motorbikes | 56.37±3.44 | 87.05±3.60 | 85.06±2.99 | 50.60±3.21 | 76.31±4.77 | **88.74±1.69** |
| | car-side | 50.21±3.16 | **98.01±3.16** | 95.34±1.65 | 51.60±7.30 | 93.24±1.51 | 91.32±2.98 |
| | dolphin | 50.38±4.61 | 61.43±4.24 | 65.86±3.34 | 52.20±3.50 | 60.84±4.42 | **69.12±1.50** |
| | goose | 57.19±4.06 | 59.69±3.98 | 63.44±3.98 | 50.60±5.03 | 50.34±4.15 | **71.81±1.84** |
| | cactus | 64.11±4.19 | 52.26±3.63 | 63.39±3.51 | 49.40±2.08 | 53.12±4.13 | **67.35±1.03** |

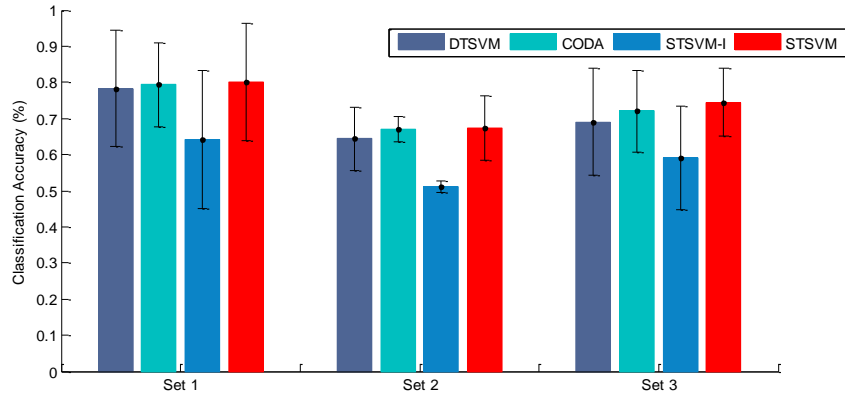

Figure 10- The average classification accuracy of each set in our approach and compared methods in the second experiment setting.

## 6 CONCLUSION

In this paper, we propose a new unified framework for self-taught learning approach. In our approach, we simultaneously learn a new space $\Im(.)$ where conditional distribution over $\Im(x)$ ( $p^t(\Im(x)|y) \approx p^s(\Im(x)|y)$ ) remains the same and learns robust SVM classifiers for the target task using both the source and target data in the new representation. The proposed approach is called Self-Taught SVM (STSVM). In our approach, the source data and its hidden labels are incorporated in STSVM objective function. To optimize our proposed objective function, an iterative approach is presented which consists of two steps. In the first step, by utilizing an iterative approach, domain adaptation is done and in the second step, the source samples are relabeled by optimizing a quadratic convex function. These steps are iterated until convergence is reached. To evaluate the proposed approach, it is applied on MSRC+LMO and Caltech-256 datasets. It should be noted that in each setting, the target and source samples can be related or unrelated. The results of our experiments show that our method, in each situation, has a superior performance to the most successful and recent existing algorithms. An important advantage of STSVM is that it transforms the samples to a domain such that conditional distribution over domains remains the same and meanwhile the discriminative knowledge of both domains is also considered. Therefore, it allows our method to achieve superior robustness in the transfer learning. It is expected that transfer learning should always improve the accuracy of the target task. However, in some situations, it leads to a decrease in performance which means negative transfer has happened. Negative transfer, in our approach, can be the consequence of invalid similarity relation between the target and source samples and noisy and incomplete labeling of target samples. In the future works, we plan to extend our approach to handle the negative transfer.

## 7 ACKNOWLEDGEMENT

This research was in part supported by a grant from the Institute for Research in Fundamental Sciences (IPM) (Grant number CS1395-4-68).

## 8 REFERENCES

[1] S. J. Pan and Q. Yang, "A survey on transfer learning," *IEEE Transactions on knowledge and data engineering,* vol. 22, pp. 1345-1359, 2010.
[2] Y. Chen, W. Guoping, and D. Shihai, "Learning with progressive transductive support vector machine.," *Pattern Recognition Letters,* vol. 24, pp. 1845-1855, 2003.

[3] A. Gammerman, V. Vovk, and V. Vapnik, "Learning by transduction," *Uncertainty in Artificial Intelligence,* pp. 148-156, 1998.

[4] J. Quionero-Candela, M. Sugiyama, A. Schwaighofer, and N. D. Lawrence, *Dataset Shift in Machine Learning*: The MIT Press, 2009.

[5] S. J. Pan and Q. Yang, "A Survey on Transfer Learning," *IEEE Transactions on Knowledge and Data Engineering,* vol. 22, pp. 1345-1359 2010.

[6] M. Sugiyama and M. Kawanabe, *Machine learning in non-stationary environments: Introduction to covariate shift adaptation*: MIT Press, 2012.

[7] F. Orabona, C. Castellini, B. Caputo, E. Fiorilla, and G. Sandini, "Model adaptation with least-squares SVM for hand prosthetics," in *ICRA*, 2009.

[8] T. Tommasi, F. Orabona, and B. Caputo, "Safety in numbers: Learning categories from few examples with multi model knowledge transfer," in *Computer Vision and Pattern Recognition (CVPR)*, pp. 3081-3088, 2010.

[9] V. H. Ablavsky, C. J. Becker, and P. Fua, "Transfer learning by sharing support vectors," No. EPFL-REPORT-1813602012.

[10] R. Raina, A. Battle, H. Lee, B. Packer, and A. Y. Ng, "Self-taught learning: transfer learning from unlabeled data," in *24th international conference on Machine learning*, pp. 759-766, 2007.

[11] H. Wang, F. Nie, and H. Huang, "Robust and Discriminative Self-Taught Learning," in *ICML*, pp. 298-306, 2013.

[12] H. DaumeIII, "Frustratingly easy domain adaptation," in *ACL*, 2007.

[13] L. Duan, I. W. Tsang, D. Xu, and S. J. Maybank, "Domain transfer svm for video concept detectio," in *IEEE Conference on CVPR*, pp. 1375-1381, 2009.

[14] S. Li, K. Li, and Y. Fu, "Self-Taught Low-Rank Coding for Visual Learning," *IEEE Transactions on Neural Networks and Learning Systems,* 2017.

[15] M. Chen, K. Q. Weinberger, and J. Blitzer, "Co-training for domain adaptation," in *Advances in neural information processing systems (NIPS)*, pp. 2456-2464, 2011.

[16] H. Shimodaira, "Improving predictive inference under covariate shift by weighting the log-likelihood function," *Journal of statistical planning and inference,* vol. 90, pp. 227-244, 2000.

[17] S. Bickel, M. Bruckner, and T. Scheffer, "Discriminative learning for differing training and test distributions," in *International conference on Machine learning*, pp. 81-88, 2007.

[18] S. Ben-David, J. Blitzer, K. Crammer, A. Kulesza, F. Pereira, and J. W. Vaughan, "A theory of learning from different domains," *Mach. Learn.,* vol. 79, pp. 151-175, 2010.

[19] P. Germain, A. Habrard, F. Laviolette, and E. Morvant, "A PAC-Bayesian Approach for Domain Adaptation with Specialization to Linear Classifiers," in *International Conference on Machine Learning*, pp. 738-746, 2013.

[20] P. Germain, A. Habrard, F. Laviolette, and E. Morvant, "A new PAC-Bayesian perspective on domain adaptation.," in *International Conference on Machine Learning*, 2016.

[21] M. Gong, K. Zhang, T. Liu, D. Tao, C. Glymour, and B. Schölkopf, "Domain adaptation with conditional transferable components," in *The 33rd International Conference on Machine Learning* pp. 2839-2848, 2016.

[22] L. Li, X. Jin, and M. Long, "Topic Correlation Analysis for Cross-Domain Text Classification," in *AAAI*, 2012.

[23] Y. Mansour, M. Mohri, and A. Rostamizadeh, "Domain adaptation: Learning bounds and algorithms," in *COLT*, 2009.

[24] J. Lu, V. Behbood, P. Hao, H. Zuo, S. Xue, and G. Zhang, "Transfer learning using computational intelligence: a survey," *Knowledge-Based Systems,* vol. 80, pp. 14-23, 2015.

[25] A. Margolis, "A literature review of domain adaptation with unlabeled data," Washington University2011.

[26] L. Bruzzone and M. Marconcini, "Domain adaptation problems: A DASVM classification technique and a circular validation strategy," *IEEE transactions on pattern analysis and machine intelligence,* vol. 32, pp. 770-787, 2010.

[27] E. Morvant, "Domain adaptation of weighted majority votes via perturbed variation-based self-labeling," *Pattern Recognition Letters,* vol. 51, pp. 37-43, 2015.
[28] F. Laviolette, M. Marchand, and J. F. Roy, "From PAC-Bayes bounds to quadratic programs for majority votes," in *International Conference on Machine Learning*, 2011.
[29] K. M. Borgwardt, A. Gretton, M. J. Rasch, H. P. Kriegel, B. Scholkopf, and A. J. Smola, "Integrating structured biological data by kernel maximum mean discrepancy," in *ISMB*, 2006.
[30] A. Rakotomamonjy, F. R. Bach, S. Canu, and Y. Grandvalet, "SimpleMKL," *Journal of Machine Learning Research,* vol. 9, pp. 2491-2521, 2008.
[31] M. Grant and S. Boyd, "CVX Users' Guide," 2012.
[32] J. Shotton, M. Johnson, and R. Cipolla, "Semantic texton forests for image categorization and segmentation," in *Computer Vision and Pattern Recognition*, pp. 1-8, 2008.
[33] C. Liu, J. Yuen, and A. Torralba, "Nonparametric Scene Parsing: Label Transfer via Dense Scene Alignment," in *Computer Vision anf Pattern Recognition*, 2009.
[34] S. Lazebnik, C. Schmid, and J. Ponce, "Beyond bags of features: Spatial pyramid matching for recognizing natural scene categories," in *Computer Vision and Pattern Recognition*, pp. 2169–2678, 2006.
[35] D. Lowe, "Distinctive image features from scale-invariant keypoints," *Int. J. Comput. Vision,* vol. 60, pp. 91–110, 2004.
[36] J. V. D. Weijer and C. Schmid, "Coloring Local Feature Extraction," in *European Conference on Computer Vision*, pp. 334-348, 2006.
[37] G. Griffin, A. Holub, and P. Perona, "Caltech 256 object category dataset," California Institue of Technology2007.
[38] A. Bosch, A. Zisserman, and X. Munoz, "Representing shape with a spatial pyramid kernel," in *ACM international conference on Image and video retrieval*, 2007.
[39] D. Lowe, "Distinctive image features from scale-invariant keypoints," *International Journal of Computer Vision,* vol. 60, pp. 91-110, 2004.
[40] O. Tuzel, F. Porikli, and P. Meer, "Human detection via classification on riemannian manifold," in *Computer Vision and Pattern Recognition*, 2007.
[41] T. Ojala, M. Pietikäinen, and T. Mäenpää, "Multiresolution gray-scale and rotation invariant texture classification with local binary pattern," *IEEE Trans. Pattern Anal. Mach. Intell. (PAMI),* pp. 971-987, 2002.